\documentclass[conference]{IEEEtran}
\IEEEoverridecommandlockouts
\usepackage{comment}
\usepackage{cite}
\usepackage{amsmath,amssymb,amsfonts}
\usepackage{algorithmic}
\usepackage{graphicx}
\usepackage{textcomp}
\usepackage{xcolor}
\usepackage{booktabs}
\usepackage{multicol} %CB
\usepackage{multirow} %CB
\usepackage{url}%CB
\usepackage[
    colorlinks=true,
    urlcolor=black,
    linkcolor=black,
    citecolor=black,
    filecolor=black,
    breaklinks=true,
    hyperfootnotes=false]{hyperref}%RSU

\def\BibTeX{{\rm B\kern-.05em{\sc i\kern-.025em b}\kern-.08em
    T\kern-.1667em\lower.7ex\hbox{E}\kern-.125emX}}

\begin{document}

\title{CAMÕES: A Comprehensive Automatic Speech Recognition Benchmark for  European Portuguese
%: A Comprehensive Benchmark for Automatic Speech Recognition in European Portuguese and Varieties \\
% Key words for the title: ASR Benchmark European Portuguese
\thanks{This work was funded by Portuguese national funds through Fundação para a Ciência e a Tecnologia (FCT), under project UIDB/50021/2020 (DOI:10.54499/UIDB/50021/2020), and by the Portuguese Recovery and Resilience Plan and NextGenerationEU European Union funds under project C644865762-00000008 (Accelerat.AI). %© 2025 IEEE. Personal use of this material is permitted. Permission from IEEE must be obtained for all other uses, in any current or future media, including reprinting/republishing this material for advertising or promotional purposes, creating new collective works, for resale or redistribution to servers or lists, or reuse of any copyrighted component of this work in other works.
}%with reference  DOI:10.54499/UIDB/50021/2020
%\thanks{This work was funded by BLIND BLIND BLIND}
\vspace{-0.7cm}
}

\author{
\IEEEauthorblockN{
Carlos Carvalho\IEEEauthorrefmark{1}\IEEEauthorrefmark{2}, 
Francisco Teixeira\IEEEauthorrefmark{1}, 
Catarina Botelho\IEEEauthorrefmark{1},
Anna Pompili\IEEEauthorrefmark{1}, 
Rubén Solera-Ureña\IEEEauthorrefmark{1}, 
Sérgio Paulo\IEEEauthorrefmark{1}, \\
Mariana Julião\IEEEauthorrefmark{1}\IEEEauthorrefmark{2},
Thomas Rolland\IEEEauthorrefmark{1},
John Mendonça\IEEEauthorrefmark{1}\IEEEauthorrefmark{2}
Diogo Pereira\IEEEauthorrefmark{1}\IEEEauthorrefmark{2}, 
Isabel Trancoso\IEEEauthorrefmark{1}\IEEEauthorrefmark{2},
Alberto Abad\IEEEauthorrefmark{1}\IEEEauthorrefmark{2}}
\IEEEauthorblockA{\IEEEauthorrefmark{1}INESC-ID, Lisbon, Portugal \IEEEauthorrefmark{2}Instituto Superior T\'{e}cnico, Universidade de Lisboa, Portugal}
\IEEEauthorblockA{}
\vspace{-1.4cm}
}

\maketitle

\begin{abstract}
%150 words
Existing resources for Automatic Speech Recognition in Portuguese are mostly focused on Brazilian Portuguese, leaving European Portuguese (EP) and other varieties under-explored. To bridge this gap, we introduce CAMÕES, the first open framework for EP and other Portuguese varieties. It consists of (1) a comprehensive evaluation benchmark, including 46h of EP test data spanning multiple domains; and (2) a collection of state-of-the-art models. For the latter, we consider multiple foundation models, evaluating their zero-shot and fine-tuned performances, as well as E-Branchformer models trained from scratch. A curated set of 425h of EP was used for both fine-tuning and training. Our results show comparable performance for EP between fine-tuned foundation models and the E-Branchformer. Furthermore, the best-performing models achieve relative improvements above 35\% WER, compared to the strongest zero-shot foundation model, establishing a new state-of-the-art for EP and other varieties.
\end{abstract}


\vspace{-0.4cm}
\section{Introduction}
\vspace{-0.1cm}
\label{introduction}
%{\color{red}Available space: {6 pages + 2 pages refs}
%Deadline: 4th of June
%}
%{\red{!! We need to always use the same form of x hours vs xh !!}}
Portuguese is the 8$^{\text{th}}$ most spoken language in the world~\cite{statista2025spoken}; it is an official language in nine countries, 
in Europe (Portugal), South America (Brazil), Africa (the so called PALOP countries: Angola, Mozambique, Guinea-Bissau, Cape Verde, São Tomé and Príncipe, and Equatorial Guinea), and Asia (East Timor). 
%spanning four continents -- Portugal, Brazil, Angola, Mozambique, Guinea-Bissau, Cape Verde, São Tomé and Príncipe, Equatorial Guinea (these 6 countries constitute the African Countries with Portuguese as Official Language (PALOP) group), East Timor -- and 
It is also spoken in regions of India (Goa) and China (Macao), %(Goa and Macao, respectively)
being the native language of about 240 million people worldwide~\cite{ethnologue_por}. Differences between varieties are mostly phonetic, phonological and prosodic, showing also lexical and syntactic variation \cite{mateus2000phonology}. %Differences between varieties are not only at the lexical and syntactic levels, but mostly at the phonetic, phonological, and prosodic levels \cite{mateus2000phonology}. %Despite this wide reach,
The variety most commonly represented in Automatic Speech Recognition (ASR) R\&D is Brazilian Portuguese (BP), which is spoken by $\sim$197 million native speakers \cite{commonvoice, candido2023coraa, fleurs2023, li2023yodas}. As a result, European Portuguese (EP) and the African and Asian Portuguese varieties (AAP) are seldom considered; few works examine these varieties independently and they are often conflated with BP \cite{candido2023coraa, spotify_portuguese}. Actually, up-to-date state-of-the-art (SOTA) ASR results for EP and AAP are non-existent, in contrast to the case of BP. %The differences between all varieties \cite{mateus2000phonology} are not only at the lexical and syntactic levels, but mostly at the phonetic, phonological, and -although less studied- the prosodic levels.
%The differences between all varieties \cite{mateus2000phonology} are not only at the lexical and syntactic level, but mostly at the phonetic and phonological levels, and although less studied, at the prosodic level.%, rouas2008
%The under-representation of EP and AAP is reflective ...

%-- despite recent advances of  --
This under-representation is reflective of broader challenges in modern supervised ASR systems which, despite recent performance improvements due to architectural advances \cite{first_speech_transformer, rnn_vs_transformers, conformer, e_branchformer, fast_conformer},
%-- despite the architectural advances of transformer-based end-to-end supervised learning approaches  that have greatly improved system performance -- 
remain heavily dependent on large-scale labelled data and require substantial computational resources to achieve strong performances \cite{chan2021speechstew, whisper, owsm, owsm_ctc}. 
%Such requirements make these approaches feasible for only a few well-resourced laboratories and companies worldwide. 
% such, although end-to-end supervised learning approaches achieve strong performances for high-resource languages such as English,
%\cite{librispeech, tedlium, people_speech_dataset, GigaSpeech2021, libriheavy}
%-- which is supported by extensive benchmarks --, 
%they remain impractical for languages with fewer resources.
Hence, building modern speech systems 
% for EP and AAP varieties 
from scratch for languages with fewer resources -- such as EP and AAP varieties -- remains a challenge~\cite{pellegrini2013corpus, carvalho21_iberspeech}. %due to the limited availability of annotated training data and research resources \cite{pellegrini2013corpus, carvalho21_iberspeech}. 
Nevertheless, given the linguistic diversity and global presence of the Portuguese language, it is imperative to work on these under-represented variants to ensure inclusive and equitable progress in real-world speech technologies. 

To bridge this gap, we introduce CAMÕES, the first comprehensive evaluation benchmark focused on EP, which also encompasses other Portuguese varieties, namely AAP and BP.
%To support model training and fine-tuning, we have organized a set of 425 hours of in-house EP speech data, and curated an evaluation set comprising 46 hours of EP data, covering a wide range of domains and demographic groups to validate the representativeness and robustness of our models. In addition, we evaluate on $\sim$9h of AAP data, and 10h of BP data.
The benchmark consists of a curated evaluation set with 46h of EP data, covering a wide range of domains and demographic groups, in addition to $\sim$3.4h and {$\sim$13.2h} of AAP and BP data, respectively. % ($\sim45.3h)$ -> with MUPE
This rich evaluation resource is used to validate the representativeness and robustness of an extensive set of speech recognition models, trained with 425h of EP speech data. 
To address the relative scarcity of labelled data, we leverage two widely adopted transfer learning strategies for low-resource settings: self-supervised learning (SSL)-based foundation models \cite{mms,zhang2023google,seamlessm4t_v2,xeus,abouelenin2025phi} and supervised foundation models \cite{whisper, owsm, owsm_ctc, owsm-v4}, achieving state-of-the-art (SOTA) results for all varieties of Portuguese in the evaluated datasets.
%, which have been shown to improve performance by transferring knowledge between languages and tasks.
%With these approaches, we 
%{\color{blue}As a result of this work that combines (1) speech data compilation, and curation, and (2) leveraging self-supervised learning (SSL)-based foundation models \cite{mms,zhang2023google,seamlessm4t_v2,xeus,abouelenin2025phi} and supervised foundation models \cite{whisper, owsm, owsm_ctc, owsm-v4} to address the relative scarcity of labelled data, we develop state-of-the-art (SOTA) models for ASR in all varieties of Portuguese. Furthermore, as far as we know, this work fills an existing gap for EP and AAP by establishing an up-to-date SOTA for ASR in this varieties, while also achieving competitive results for BP.} Overall, our contributions can be summarized as follows:
%
%These models typically fall into two main broad categories: . 
%In this work, we leverage both approaches to develop state-of-the-art models for European Portuguese, and African and Asian Portuguese varieties !!Brazilian also??.
%
Overall, our contributions can be summarized as follows:
\begin{enumerate}
    \item We introduce the first comprehensive and publicly available ASR benchmark for EP and other Portuguese varieties, designed to foster research 
    %and development 
    in this language;
    \item We evaluate both zero-shot and fine-tuned performance of a range of foundation models, including speech-centric models such as Whisper Large v3 (WhisperLv3) \cite{whisper}, OWSM-CTC v4 \cite{owsm_ctc}, Massively Multilingual Speech (MMS)-all \cite{mms} and SeamlessM4T-v2 \cite{seamlessm4t_v2}; as well as multimodal large language models (LLMs) such as Phi-4-Multimodal Instruct (Phi-4-MI) \cite{abouelenin2025phi};%, incorporating different prompting strategies;
    \item We train E-Branchformer (EBranch) \cite{e_branchformer} models from scratch, without and with SSL features (EBranch-SSL);
    \item 
   We develop state-of-the-art models for ASR in all varieties of Portuguese and release them on Hugging Face;%\footnote{%We intend to make this benchmark available through a HuggingFace LeaderBoard  after the anonymous review period, together with the models.
%\url{https://huggingface.co/inesc-id}};%{\color{red}We obtain SOTA ASR models for all Portuguese varieties and release them on Hugging Face}
    \item 
    We fill an existing gap in ASR R\&D for EP and AAP by establishing up-to-date SOTA performance references for these two varieties.
    %Furthermore, as far as we know, this work fills an existing gap for EP and AAP by establishing an up-to-date SOTA for ASR in this varieties, while also achieving competitive results for BP
   %\footnote{These models will be released following the anonymous review period.}
\end{enumerate} %(2) (3) 
% Add comment about how the model will be made publicly available?

%The remainder of this work is organized as follows: Section \ref{sec:related} presents related work; Section \ref{sec:data} describes the data used in this work; Section \ref{sec:camoes} introduces the proposed benchmark and models; Section \ref{sec:exp} contains details about experimental parameters; Section \ref{sec:results} describes our results; and finally, Section \ref{sec:conclusions} presents the main conclusions of this work.

\vspace{-0.1cm}
\section{Related Work}
\vspace{-0.1cm}
\label{sec:related}

\subsection{Benchmarking ASR Models in Low Resource Scenarios}
ASR model benchmarking is important not only for an in-depth understanding of the strengths and limitations of the multiple architectures and training procedures, but also for raising community awareness about datasets and tools
%, common assessment metrics or, even, evaluation and pre- and post-processing scripts 
available for a given task or language. For instance,
%SUPERB \cite{s3prl} provides a comprehensive assessment benchmark of SSL models applied to ASR, among other speech processing tasks. 
%For instance, SUPERB \cite{s3prl} is a leaderboard to benchmark the performance of SSL models. It provides a comprehensive assessment of SSL models applied to automatic speech recognition, among other speech processing tasks. 
ML-SUPERB \cite{Shi2023MLSUPERBMS} (a multilingual extension of SUPERB\cite{s3prl}) provides a comprehensive assessment benchmark of SSL models applied to ASR 
%presents a multilingual extension of SUPERB 
that covers 143 languages, ranging from high-resource to endangered. %, and report results showing speech SSL models can significantly improve performance compared to FBANK features.
In the context of low-resource languages, large-scale multilingual ASR systems \cite{whisper, seamlessm4t_v2, mms} have been adopted with some success
%, showing cross-lingual speech representation can have a positive impact on systems developed 
%in resource-scarce contexts 
for languages like Urdu\cite{URDU-BENCH}, Thai\cite{THAI_BENCH} or Greek varieties\cite{greek_benchmarking}. Such models have been assessed in zero-shot (Whisper \cite{pashto_punjabi_urdu-benchmarking, greek_benchmarking, THAI_BENCH, URDU-BENCH}, XLSR-Wav2Vec2 \cite{greek_benchmarking}, MMS and Seamless-M4T \cite{URDU-BENCH}) and fine-tuning \cite{pashto_punjabi_urdu-benchmarking, THAI_BENCH, URDU-BENCH} scenarios. 
%However, although fine-tuning generally helps to improve model performance, most of the systems show an increased gap between human and automatic performance on out-of-domain data.
While fine-tuning generally improves performance, most systems still show relatively low recognition accuracy on out-of-domain data.

\vspace{-0.1cm}
\subsection{ASR research for Portuguese}
\vspace{-0.1cm}
Research on ASR for Portuguese dates back to the late 1990's  \cite{ASR_PT_FIRST,ASR_FIRST_BR}. While EP was the initial focus, the shift to resource-intensive end-to-end (E2E) speech recognition has increasingly favoured BP, driven by its much larger population and the resulting availability of more extensive datasets. In contrast, ASR for EP could not benefit from these advances. 
% in end-to-end architectures and the availability of corpora 
 %that supported progress in BP.% mostly due to the lack of resources. % and other resource intensive languages.

\subsubsection{European Portuguese}
%In early 2000's, automatic transcription of broadcast news (BN) became a mainstream research topic. Thus, the AUDIMUS system~\cite{AUDIMUS_BN}, and its variants~\cite{NetoASR2008}, were the first attempts to automatically transcribe such contents in EP.

%In early 2000's, automatic transcription of  became a mainstream research topic. Thus, 
The AUDIMUS system~\cite{AUDIMUS_BN,NetoASR2008} was among
%and its variants~\cite{NetoASR2008}, was
the first attempts to automatically transcribe broadcast news (BN) in EP, leveraging a hybrid HMM/MLP approach.
%, being later incorporated into a fully-automatic live subtitling solution~\cite{NetoASR2008}.
%More recently, works adopting end-to-end approaches for EP
%in the 
%constrained-resource 
%EP scenario 
%have shown considerable limitations.
The hybrid HMM/DNN framework remained SOTA for a long period, even after the emergence of E2E models. Recent work with E2E models is scarce and evidences a lack of resources. An early CTC-attention E2E model trained from scratch using $\sim$180h obtained WERs two to three times worse than a hybrid HMM/DNN system~\cite{carvalho21_iberspeech}. Other works attempted to leverage large pre-trained English models. In ~\cite{MouraodeSa2021mscthesis}, models fine-tuned with EP data were outperformed by a hybrid baseline, whereas modest improvements for telephone speech were achieved in~\cite{Medeiros2023thesis,Medeiros2023futureinternet} using different mixtures of EP and BP speech for fine-tuning.

\subsubsection{Brazilian and other Portuguese Varieties}
%The first attempts to build large-vocabulary ASR systems for other Portuguese varieties followed those of EP~\cite{Silva05ASR-BR,NelsonNeto_ASR_BR}. 
The first attempts to build large-vocabulary ASR systems for other Portuguese varieties appeared soon after those for EP~\cite{Silva05ASR-BR,NelsonNeto_ASR_BR}, with early works focusing on adapting existing EP models to the specificities of BP~\cite{abad2009interspeech} and African Portuguese (AP)~\cite{Koller2010interspeech} varieties, and on the development of multi-variety setups through automatic accent identification~\cite{abad2012propor}.
%\footnote{Despite the differences of the distinct AP varieties, they have been usually considered as a single broad variety in research given the difficulty to collect enough data from each country, and the difficulty to distinguish varieties in the more uniform BN speech (which has been a main resource of work).}~\cite{Koller2010interspeech} varieties.
%, addressing the collection of new corpora, the analysis of the main sources of variability, and the development of technical solutions,
%at the lexical, acoustic, and syntactic levels, 
%leading to remarkable performance improvements. 
%The automatic identification of Portuguese varieties (EP/BP/AP) in a multi-variety speech recognition setup was addressed in~\cite{abad2012propor}.
%
In contrast to the EP case, the release of several large corpora specifically targeting BP \cite{alencar2008lsf, candido2023coraa,limaintelligentsystems2025,leal2025mupe}
%(\kern-0.205em\cite{alencar2008lsf, candido2023coraa,limaintelligentsystems2025,leal2025mupe})
%{\color{red}(CETUC \cite{alencar2008lsf}, CORAA \cite{candido2023coraa}, NURC-SP \cite{limaintelligentsystems2025}, MuPe Life Stories \cite{leal2025mupe})} 
has fostered research for this variety over the past decade. After hybrid systems~\cite{BR_KALDI}, different E2E approaches based on pre-trained SSL architectures were developed for BP in \cite{grisCPPL2022,candido2023coraa,limaintelligentsystems2025,leal2025mupe}, using different datasets, fine-tuning strategies, and data augmentation methods~\cite{LLM_dataAugment_BR}, achieving significant improvements. 
%Recent works explore LLM text generation, text-to-speech and voice conversion for BP data augmentation leading to significant improvements~\cite{LLM_dataAugment_BR}.
To the best of our knowledge, there are no noteworthy contributions on ASR for Asian Portuguese, likely due to a lack of data resources.

\vspace{-0.05cm}
\section{Data resources and preparation}
\vspace{-0.05cm}
\label{sec:data}
%This section provides extensive information on the corpora used in the experimental evaluation and the data curation process underlying it. %CARLOS: mention here the name of the benchmark 
\begin{table*}[t]%{p{1cm} p{1cm} p{1cm} p{1cm} p{1cm} p{1cm} p{1cm} p{1cm} p{1cm} p{1cm} p{10cm}}
\setlength{\tabcolsep}{4pt}
\caption{CAMÕES Benchmark: Train and test partition statistics per domain.
%Corpora marked with * cannot be made available. **Indicates that speakers are shared among train and test partitions. -- Indicates dataset is not used in the specific partition. 
%\textbf{RS}: Read Speech, \textbf{BN}: Broadcast News, \textbf{T/L}: Talk/Lectures, \textbf{TV}: Tv show, entertainment, \textbf{SI}: Sociolinguistic Interviews, 
\textbf{M$\vert$F}: percentage of male and female speakers in the dataset -- the total may not be 100\% due to speakers with unknown gender, \textbf{NI} indicates information not available.} 
\vspace{-0.2cm}
\centering
\resizebox{0.9\textwidth}{!}{
%\scalebox{0.90}{
\label{table:corpora}
\begin{tabular}{cl cccc c c p{8cm}}
\toprule
& & \multicolumn{2}{c}{\bf Train} & \multicolumn{2}{c}{ \bf Test} &  & & \\ \cmidrule{3-4} \cmidrule{5-6}
\bf Domain & \bf Corpus & \bf Hrs & \bf \#Spks & \bf Hrs & \bf \#Spks & \bf Age & \bf M$|$F (\%) & \multicolumn{1}{c}{\bf Notes}   \\
\midrule
\multirow{10}{*}[0pt]{\bf RS} & BD-Publico~\cite{neto1997design}  & 21.8 & 100 & 2.0 & 10 & 18--28 & 50$|$50 & 
%Created in 1997, comprises r
Read sentences extracted from an EP newspaper. \\% PÚBLICO. \\
& CommonVoice~\cite{commonvoice} & -- & -- & 1.8 & 42 & 13--59 & 48$|$12 & Speaker count estimated from the client ids provided in the corpus.\\
& DIRHA~\cite{dirha} & 2.2 & 20 &  -- & -- & 20--60 & 50$|$50 & Read and spontaneous home automation commands.\\
& HLT TTS~\cite{spauloPhD} & 68.3 & 20 &  -- & -- & 13--64 & 60$|$40 & In-house dataset recorded for TTS training. \\ %8 speakers without demographic information. \\
& MLS\_extended & 54.8 & 12 & 1.0 & 10 & NI & 27$|$73 & EP extension of MLS~\cite{mls}: automatically aligned audiobooks. \\
& PT\_Adults~\cite{pt_adults} & 7.3 & 66 & 1.6 & 17 & 25--59 & 52$|$48 & Corresponds to YMA in~\cite{pt_adults}.\\
& PT\_Children~\cite{pt_children} & -- & -- & 2.1 & 52 & 3--10 & 56$|$44 & Corpus of child   speech. \\ %Test speakers are aged between ??--??. \\
%& PT\_Children$_{\text{ag}2}$~\cite{} & -- & -- & 2.5 & \color{red} ?? & \\
& PT\_Elderly~\cite{pt_elderly} & 48.2 & 794 & 1.3 & 172 & 60-100 & 26$|$74 & Train/test speakers are aged between 60-75/76-100 years, except for 55 speakers in train with an unknown age $<$59. \\
& SpeechDat~\cite{hagen2003hmm} & 30.2 & 3,349 & 9.7 & 604 & 14--98 & 46$|$54 & Telephone speech sampled at 8kHz, upsampled to 16kHz. \\ 
%A collection of speech read from telephone calls, collected by a  Portuguese telecommunications operator. \\
%\cmidrule{2-6}
%& \multicolumn{1}{c}{\textbf{Subtotal}} & \bf 232.8 & \bf 4,361 & \bf 19.4 & \bf 907 & & &  \\ 

\midrule
\bf BN & Alert~\cite{trancoso2003evaluation} & 45.6 & 1,356 & 6.6 & 175 & NI & 70$|$29 & Broadcast news data. \\
%Created in cooperation with a Portuguese public service broadcasting organization. \\ %in 2000-2001.\\
\midrule
\multirow{3}{*}[0pt]{\bf T/L} & CORAA~\cite{candido2023coraa} & 2.2 & 183 &  -- & -- & NI & NI & European Portuguese subset. \\
& Lectra~\cite{trancoso2008lectra} & 22.0 & 7 & 2.6 & 7 & NI & NI & 
%A corpus of European Portuguese 
University lectures. Speakers are shared among partitions. \\ % maybe we don't mention this because we are not consistent over datasets (pós acordo)? --> all the test set has been converted. If Carlos is using it, may be the best place to mention this part is in the "text pre-processing" section.
& MuAViC~\cite{muavic} & 19.2 & 100 & 0.4 & 2 & NI & 60$|$40 & TEDx talks. \\
%\cmidrule{2-6}
%& \multicolumn{1}{c}{\textbf{Subtotal}} & \bf 43.4 & \bf 290 & \bf 3.0 & \bf 9 & & &  \\ 

\midrule
 \multirow{3}{*}[0pt]{\bf CS} &  
 Coral~\cite{trancoso1998corpus}  & 6.0 & 28 &  -- & -- & 19--29 & 50$|$50 & Map task dialogues. \\% Assigned to TV, due to the degree of spontaneity. %This corpus corresponds to dialogues for executing a map task. It is assigned to the TV category, to match the degree of spontaneity and dialogue style.\\
& Postport~\cite{meinedo2010l2f} & 31.3 &  {\scriptsize $>$}247 & 3.9 & {\scriptsize $>$}30 & NI & 54$|$24 &  
%A corpus collected to analyse and evaluate European Portuguese and its variation. 
Debates and entertainment (a few  documentaries and  information). \\
& VoxCelebPT~\cite{voxcelebpt} &  -- & -- & 2.9 & 13 & NI & 38$|$62 & Voices of Portuguese celebrities collected from YouTube. \\
%\cmidrule{2-6}
%& \multicolumn{1}{c}{\textbf{Subtotal}} & \bf 37.3 & \bf 286 & \bf 6.8 & \bf 45 & & &  \\ 
\midrule

 \multirow{2}{*}[0pt]{\bf SI} & Fala Bracarense~\cite{falabracarense} & 66.1 & 75 & 6.1 & 8 & 15--92 & 45$|$55 & Recorded in the city of Braga, collected between 2009-2014.%, a city in northern Portugal.
\\
& PT Fundamental~\cite{portuguesfundamental} &  -- & -- & 4.2 & 169 & 17--69 & 44$|$56 & Low quality recordings of interviews collected in the 1970's. \\
%\cmidrule{2-6}
%& \multicolumn{1}{c}{\textbf{Subtotal}} & \bf 66.1 & \bf 75 & \bf 10.3 & \bf 177 & & &  \\ 

\midrule
%& \multicolumn{1}{c}{\bf Total} & 425.1 & 6,368 & 48.1 & 1,319 \\
& \multicolumn{1}{c}{\bf Total} & \bf 425.2 & \bf 6,357 & \bf 46.2 & \bf 1,311 \\
\bottomrule
\end{tabular}}
%}
\vspace{-0.4cm}
\end{table*}
\subsection{European Portuguese Corpora}
%\label{sec:data}
% CB: I would avoid saying collected, as that indicates that we actually collcted the data
%To benchmark, train, and fine-tune different models, a set of 19 corpora has been collected. 
We curated a set of 18 corpora to train, fine-tune and benchmark the different EP ASR models of this work: 14 for training and fine-tuning, and 14 for the evaluation benchmark. These resources correspond to a mix of proprietary corpora collected over the years through various research collaborations, corpora recorded in-house, and corpora crawled from publicly available online sources.
Ground truth transcriptions were obtained using different methods:
%depending on the type of resource: 
TV shows and broadcast news were  manually annotated; spoken books were automatically aligned with their source text at a book or chapter level; for other read speech corpora, the original prompt was used as a reference. 

Table \ref{table:corpora} provides a brief description of the subsets of each corpora, along with key statistics, including duration, number of speakers, gender, and age information. In total, we gathered over 470h of data, from which we created a training set comprising 425h (denoted as EP-425), and a test set of 46h for benchmarking. The information in Table \ref{table:corpora} represents our own curated version of each dataset, corresponding to clean and partitioned subsets 
instead of the original datasets. %FOOTNOTE}
%of the full datasets. %, obtained after cleaning and partitioning the datasets. FOOTNOTE}

%Transcriptions were performed manually. %CB: my suggestion:
%Ground truth transcriptions consist of manual transcriptions, prompts or books that were automatically aligned.

%!!TODO: Mention that we are not necessarily using the whole corpora, but sometimes a subset we curated -- done, check above
\vspace{-0.02cm}
%\subsection{Corpora for Brazilian and AAP Portuguese Varieties}
\subsection{Corpora for Brazilian, African and Asian Portuguese Varieties}
\vspace{-0.03cm}
\label{sec:data_BR}
For the experiments with Portuguese varieties other than EP, we used
%four
four corpora: PoSTPort~\cite{rouas2008portuguese} and Português Falado~\cite{bettencourt2000portugues} for AAP, and CETUC~\cite{alencar2008lsf}, CORAA~\cite{candido2023coraa} and Português Falado~\cite{bettencourt2000portugues} for BP, which are summarized in Table~\ref{tab:baap}. For AAP, the PoSTPort corpus was used for training ($\sim$8h, denoted as AAP-8), and Português Falado for evaluation ($\sim$3.4h, excluding BP data).
% IN CASE OF MUPE -> comment below and uncomment line after
For BP, CETUC and the training partition of CORAA were used for training ($\sim$417h, denoted as BP-417), and Português Falado BP subset and CORAA's test set were used for evaluation ({$\sim$13.2h}). Additionally, MuPe's test set~\cite{leal2025mupe}, with 32.9h, was used only for comparison with the SOTA for BP.
%For BP, CETUC and the training partition of CORAA are used for training ($\sim$417h, denoted as BP-417), and Português Falado BP subset,  CORAA's test set and MuPe's test set are used for evaluation ($\sim$45.3h).

% Português falado is also called Spoken Portuguese in the elra website

\begin{table}[t]
\setlength{\tabcolsep}{2pt}
\caption{Corpora of non-European varieties.}
\vspace{-6pt}
\centering
%\resizebox{0.35\textwidth}{!}{
\resizebox{0.45\textwidth}{!}{
\label{tab:baap}
%\begin{tabular}{lcccc}
\begin{tabular}{lcccccc}
\toprule
& & \multicolumn{2}{c}{\bf Train} & \multicolumn{2}{c}{ \bf Test} &  \\
% \textbf{Corpus}  & \textbf{Varieties} & \textbf{Hrs} & \textbf{\# Spks} & \textbf{Domain} \\   \midrule
\cmidrule{3-6}
\textbf{Corpus}  & \textbf{Varieties} & \textbf{Hrs} & \textbf{\#Spks} & \textbf{Hrs} & \textbf{\#Spks} & \textbf{Domain} \\   \midrule
PosTPort~\cite{rouas2008portuguese}  & AAP & 8.2 & 384 & -- & -- & CS \\ %Mix      \\
%Português Falado~\cite{bettencourt2000portugues} & AAP/BP     & 8.7                   & --    & Mix \\ 
Português Falado~\cite{bettencourt2000portugues} & AAP & -- & -- & 3.4  &  50  & CS,SI \\ %Mix \\ 
\midrule
CETUC~\cite{alencar2008lsf}  & BP   & 145  & 100  & -- & -- & RS \\
CORAA~\cite{candido2023coraa} & BP & 272  & 1,131  & 11.2 & 58 & T/L,CS \\
%CORAA-train~\cite{candido2023coraa}      & BP & 272                     & 1,131  & & & T/L,CS \\ %Mix           \\
%CORAA-test~\cite{candido2023coraa}       & BP  & & & 10.4                    & 44                   & T/L,CS \\ %Mix  \\
%MuPe~\cite{leal2025mupe}             & BP  & -- & -- &  32.9 &   35 & SI \\ %Spontaneous \\
Português Falado~\cite{bettencourt2000portugues} & BP & -- & -- & 2.0 &   51    & CS, SI \\ %Mix \\ 
\bottomrule
\end{tabular}}
\vspace{-16pt}
\end{table}

\vspace{-0.05cm}
\section{CAMÕES}
\label{sec:camoes}
\vspace{-0.05cm}
\subsection{Evaluation Benchmark}
\label{CAMOES_benchmark}
%\vspace{-0.05cm}
To create the CAMÕES benchmark, we carefully curated the resources described in Section \ref{sec:data} to obtain a diverse evaluation set in terms of type of speech and speaker demographics. As shown in Table \ref{table:corpora}, the corpora span a range of domains -- from read speech, 
%which is typically clean, 
%to prepared speech, such as that found in broadcast news or university lectures, and 
to more challenging conversational speech, such that in TV shows or everyday interactions. Furthermore, these corpora comprise different age groups -- children, adults, and the elderly -- as well as different regional varieties of EP. 
To ensure the representativeness of our benchmark, we organized the available test data in five domains according to the level of spontaneity (from lower to higher):
 
\textbf{Read Speech (RS)}: %Speech corresponding to read audiobooks and other read text prompts, 
Read audiobooks and text prompts, such as news articles, speech commands, numbers, single words and digits, with little to no spontaneity.

\textbf{Broadcast News (BN)}: News content from public Portuguese TV channels, chosen as an individual domain due to its particularities, i.e., mostly read or planned speech with a specific type of enunciation, uttered by professionals.
%to the particularities of this type of speech 

\textbf{Talks/Lectures (T/L)}: %Speech from 
TEDx talks and university lectures; this type of speech is prepared but not read, with a higher degree of spontaneity than previous domains. %university lectures are more spontaneous and have lower quality recordings than TEDx talks.

\textbf{Conversational Speech (CS)}: 
%\textbf{Television shows and entertainment (TV)}: 
%Speech from various sources, including 
Celebrity interviews, map task dialogues, and 
other recordings
%conversational speech 
from Portuguese TV channels. %featuring conversational speech (e.g. TV shows and political debates). 
This domain is characterized by spontaneous and interactive dialogues including more informal and demanding speech settings than previous domains.
%speech from various sources, including interviews with Portuguese celebrities, goal-oriented dyadic discourse (map task), and content from Portuguese TV channels featuring conversational speech (e.g. TV shows and political debates). This domain is characterized by spontaneous and interactive dialogues including more informal and demanding speech settings than the previous domains.

\textbf{Sociolinguistic Interviews (SI)}: 
Highly spontaneous conversational speech,
%guided by a topic
recorded in various Portuguese regions and social contexts, 
%These conversations are highly spontaneous and 
often with poor recording conditions and highly accented speech, making this the most challenging domain in this benchmark. % considered in this work.

% The corpora from which we obtain the test sets for each of the domains mentioned above are annotated in Table \ref{table:corpora}, and collectively correspond to the CAMÕES benchmark.
%Even though each domain is not equally represented, in this work we perform per-domain averages over the obtained results, such that each domain is equally represented in the final score. 
%\red{!!! note about demographics? !!!}

Some corpora may span multiple domains. In such cases, we assign the corpus to the domain that represents the predominant portion of the data.
Despite all domains not being equally represented, 
% both in the benchmark and in this work 
we report per-domain performance averages, so that all domains contribute equally to the final score.
A leaderboard for this benchmark and trained models are available in Hugging Face\footnote{%We intend to make this benchmark available through a HuggingFace LeaderBoard  after the anonymous review period, together with the models.
\href{https://huggingface.co/datasets/inesc-id/camoes_asr}{https://huggingface.co/datasets/inesc-id/camoes\_asr}}. %
%over the obtained results 

%In this way, it serves as a valuable resource for the speech research community, providing corpora with increasing levels of difficulty, from read-speech to real-world spoken data.

%A public benchmark and leaderboard is available at \red{\url{colocar.url.pt}}

\subsection{Automatic Speech Recognition models}
\label{CAMOES_framework}

In addition to establishing an evaluation benchmark, our goal is to develop SOTA ASR models for all Portuguese varieties. To this end, we leverage foundation models spanning two transfer learning paradigms: (1) supervised and (2) self-supervised learning (SSL), which are described below. %To evaluate CAMÕES, we leverage SOTA foundation models spanning two transfer learning paradigms: (1) supervised and (2) self-supervised learning (SSL), as outlined in Section~\ref{introduction}. %Below, we detail all considered models by category.   

\subsubsection{Supervised foundation models}
%Supervised foundation models are large-scale models trained on vast corpora, either through a two-step process—initial SSL on unlabelled data followed by fine-tuning on labelled data—or solely using labelled data. 
%The models discussed in this section were chosen for their potential (i.e., Phi-4-MI) or demonstrated effectiveness in adapting to low-resource languages~\cite{owsm-v4, xeus}.

\textbf{MMS-all} \cite{mms} is a large-scale ASR model based on the wav2vec 2.0 architecture \cite{baevski2020wav2vec}, with $\sim$1B parameters. It was pre-trained on 491kh of multilingual speech and fine-tuned with 107kh of labelled data from more than 1,000 languages. For Portuguese, at least $\sim$285h of labelled data were used -- details regarding the Portuguese subset were not disclosed.
%of the MMS-lab dataset are undisclosed~\cite{mms}.
%excluding any potential contribution from the MMS-lab dataset.%, for which no details have been disclosed regarding its Portuguese subset.

%\textbf{MMS-all} \cite{mms} is a large-scale multilingual model developed by Meta, built on the wav2vec-2.0 \cite{baevski2020wav2vec}  architecture with  $\sim$1 billion parameters. The model was first pre-trained in a self-supervised manner with 491k hours of multilingual speech data covering over 1,000 languages, and was subsequently fine-tuned for ASR across 1,162 languages with 107K hours of supervised data. For Portuguese, the minimum amount of supervised training data used in MMS-all is approximately 285 hours. This estimate excludes additional hours potentially contributed by the MMS-lab dataset \cite{mms} -- which is based on New Testament readings in various languages -- as detailed information about its Portuguese subset has not been disclosed. 
\textbf{SeamlessM4T-v2} \cite{seamlessm4t_v2} is a 2.31B parameter multilingual, multimodal translation model -- comprising a speech conformer encoder and a transformer text encoder-decoder -- 
that supports
%ASR and other tasks in 
101 languages. 
%It is composed by w2v-bert-2.0 (a conformer speech encoder), a transformer text encoder-decoder, and a non-autoregressive unit decoder. 
It was trained on 406kh of aligned data from the SeamlessAlign corpus \cite{seamlessm4t_v2}; the model’s data coverage for Portuguese is unspecified.

%\textbf{SeamlessM4T-v2} \cite{seamlessm4t_v2} is a 2.31 billion parameters foundational, massively multilingual and multimodal machine translation model developed by Meta. A key feature of this model is its ability to handle a wide range of speech and text tasks, including ASR, for 101 languages. SeamlessM4T consists of several components, including w2v-bert-2.0, a conformer speech encoder mentioned above, a transformer text encoder-decoder, and a transformer encoder with a non-autoregressive decoder for text-to-unit translation. The model was trained on 406k hours of automatically aligned data using the SeamlessAlign corpus \cite{seamlessm4t_v2}. However, there is no specific information available regarding the amount of speech recognition data used for Portuguese. 

\textbf{WhisperLv3}~\cite{whisper} is a 1.55B parameter transformer encoder-decoder model trained on 5Mh, covering $\sim$100 languages.
%-- with a vast majority of English. 
At least 9kh of Portuguese data were used, with no details available about the specific variety.
Despite its strong performance, it has been shown to struggle under noisy conditions, which can be alleviated applying voice activity detection (VAD) using WhisperX (hereafter WhisperLv3-X)~\cite{whisperx}.

\textbf{OWSM} \cite{owsm_ctc,owsm-v4} is an open-source initiative aimed at replicating Whisper’s performance.
%using only public data and open-source tools. 
We use OWSM-CTC v4 \cite{owsm-v4}, a 1.01B parameter encoder-only model trained on 
%a large collection of public speech data, with YODAS \cite{li2023yodas} as a key addition to the training set.
%, a large-scale web-crawled dataset licensed under Creative Commons
%The integration of YODAS improved multilingual benchmark performance over earlier OWSM versions~\cite{owsm-v4}. 
%The training dataset includes 
290kh of speech across 151 languages, with 10.8kh in Portuguese.

\textbf{Phi-4-MI} \cite{abouelenin2025phi} is a 5.57B parameter multimodal LLM with integrated speech capabilities. It was pre-trained on 2.3Mh of speech-text pairs across 8 languages, with no further details available.
%Similar to WhisperLv3, the training data is not public.
%
Phi-4-MI supports textual prompts for ASR, allowing flexible context-aware decoding.
%Whereas Whisper offers similar prompting, Phi-4-MI's LLM decoder may better utilize context, particularly in zero-shot scenarios.

\subsubsection{SSL foundation models} 
%SSL foundation models are large models trained on vast amounts of unlabelled data, where the input itself provides the supervision signal for the loss \cite{mohamed2022self}. These models
%SSL models can serve as very informative feature extractors for ASR models.

\textbf{E-Branchformer}~\cite{e_branchformer} (EBranch) is used in our from scratch training experiments, first as a baseline using FBank features, and then by incorporating SSL-based
%speech foundational 
encoders as feature extractors. %-- to isolate the impact of large-scale self-supervised pre-training.
We use XLSR and w2v-BERT2 (EBranch-XLSR and EBranch-w2vBERT2, respectively).

\textbf{XLSR} \cite{babu2021xls} is a large-scale foundation model for cross-lingual speech representation learning, based on the wav2vec 2.0 architecture. 
%At the time of its release, it marked the largest effort to democratize speech technology using only publicly available data. 
The largest variant, used in this work, has 2B parameters and was trained using SSL on 436kh of unlabelled speech from 128 languages, 17.8kh in Portuguese, mostly corresponding to EP~\cite{wang2021voxpopuli}. 

%For Portuguese, 17.8k hours were included, with nearly all (17.5k hours) sourced from VoxPopuli, an unlabelled dataset of European Parliament recordings \cite{wang2021voxpopuli}.
%\textbf{XLSR} \cite{conneau2020unsupervised} is a large-scale foundation model for learning cross-lingual speech representations based on the wav2vec 2.0 architecture \cite{baevski2020wav2vec}, representing the largest effort at the time of its proposal to make speech technology accessible for many languages using only publicly available data. The largest available model, which is used in this work, contains approximately 2 billion parameters. It was trained on roughly 436k hours of unannotated speech data covering 128 languages. Of this data, about 17.8k hours are Portuguese, with nearly all (17.5k) hours being derived from VoxPopuli, i.e. recordings of European Parliament proceedings\cite{wang2021voxpopuli}.

%CARLOS: 17.5k from VoxPopuli, 284.59h from MLS corpus, 30h from Common Voice and Voxlingua107 with 64h 
%
\textbf{w2v-BERT2} is the 600M parameter speech encoder module used in SeamlessM4T-v2 \cite{seamlessm4t_v2}. It features a conformer-based architecture \cite{conformer} pre-trained using the w2v-BERT2 algorithm \cite{chung2021w2v} with 4.5Mh of audio data. Although data was collected from publicly available sources, information on language distribution has not been disclosed. 

%CARLOS: no info about pre-training data: closed-data model
%\begin{table*}[ht]
%\setlength{\tabcolsep}{4pt}
%\caption{WER [\%] for zero-shot foundation models on the CAMÕES benchmark. §The 491k and 4.5M hours of unlabelled speech for pre-training the MMS-all and SeamlessM4T-v2 speech encoders are not included here.}
%\centering
%%\scalebox{0.98}{
%\label{tab:zero-shot-results}
%\begin{tabular}{lccc|ccccc||c |c|c| c}
%\toprule
%\multicolumn{4}{c}{} & \multicolumn{5}{c}{\bf EP} & & \multicolumn{1}{c}{\bf AAP} & \multicolumn{1}{|c|}{\bf BP} & \\
%\cmidrule{5-12}
%\multicolumn{1}{c}{\bf Model}  & \bf \#Pr. & \bf Data(h) & \bf \#Lang. &\bf RS & \bf BN & \bf T/L & \bf TV & \bf SI & \bf Avg. & \bf Avg. & \bf Avg.\\
%\toprule
%MMS-all & 0.97B & 107K\textsuperscript{§} & 1162 & 33.9 & 25.7 & 40.3 & 38.2 & 66.3 & 40.9 & 57.8 & 51.4 \\
%OWSM-CTC v4 & 1.01B & 290k & 151 & 22.5 & 24.4 & 32.0 & 28.7 & 53.3 & 32.2 & 37.7 & 33.7 \\
%Phi-4-MI & 5.57B & 2.3M & 8 & \bf 15.5 & 8.6 & 17.9 & 21.9 & \underline{45.6} & \underline{21.9} & \bf 27.8 & 29.1 \\
%SeamlessM4T-v2 & 2.31B & 406K\textsuperscript{§} & 101 & 26.7 & 17.3 & 26.3 & 27.9 & 65.3 & 32.7 & 46.9 & 34.9 \\
%Whisper Large v3 & 1.55B & 5M & 100 & 32.4 & \bf 7.9 & \bf 15.4 & \underline{18.3} & 49.74 & 24.8 & 30.5 & \underline{26.4}\\
%WhisperX Large v3 & 1.55B & 5M & 100 & \underline{16.5} & \underline{8.2} & \underline{16.6} & \bf 15.3 & \bf 40.9 & \bf 19.5 & \underline{29.7} & \bf 26.1 \\ 
%\bottomrule
%\end{tabular}
%\centering
%\end{table*}
\begin{table*}[ht]
\caption{WER [\%] on the CAMÕES benchmark. ()$\rightarrow$ denotes pre-training + finetuning on the specified datasets. Bold = best, underline = second-best.\textsuperscript{¶} Trainable parameters only — frozen: 580.49M (w2v-BERT2), 2.17B (XLSR).}
\vspace{-0.1cm}
\centering
\label{tab:results-all}
\resizebox{0.90\textwidth}{!}{
\begin{tabular}{c lcc|ccccc|c||c|c}
\toprule
& \multicolumn{1}{c}{\multirow{2}{*}[-0.3em]{\textbf{Model}}} & \multirow{2}{*}[-0.3em]{\textbf{\#Trainable Parameters}} & \multirow{2}{*}[-0.3em]{\textbf{Training Data}}
& \multicolumn{6}{c||}{\textbf{EP}}
& \multirow{2}{*}[-0.3em]{\textbf{AAP}} & \multirow{2}{*}[-0.3em]{\textbf{BP}} \\
\cmidrule{5-10}
& & & & \textbf{RS} & \textbf{BN} & \textbf{T/L} & \textbf{CS} & \textbf{SI} & \textbf{Avg.} & & \\ \midrule
\multirow{6}{*}[0pt]{0-shot} & MMS-all & - & - & 33.9 & 25.7 & 40.3 & 38.2 & 65.4 & 40.7 & 57.5 & 50.5 \\
& OWSM-CTC v4 & - & - & 22.5 & 24.4 & 32.0 & 28.7 & 52.1 & 31.9 & 37.1 & 32.3 \\
& Phi-4-MI & - & - & 15.5 & 8.6 & 17.9 & 21.9 & 44.5 & 21.7 & 27.1 & 25.9 \\
& SeamlessM4T-v2 & - & - & 26.7 & 17.3 & 26.3 & 27.9 & 64.5 & 32.5 & 46.4 & 33.3 \\
& WhisperLv3 & - & - & 32.4 & 7.9 & 15.4 & 18.3 & 49.0 & 24.6 & 29.9 & 25.8 \\
& WhisperLv3-X & - & - & 16.4 & 8.2 & 16.6 & 15.3 & 39.3 & 19.2 & 29.0 & 24.6 \\
\midrule
\multirow{3}{*}[0pt]{FT} & Phi-4-MI & 1.3B & EP-425 & 9.6 & 7.2 & 16.7 & 24.4 & 59.5 & 23.5 & 35.6 & 31.1 \\
& WhisperLv3 & 1.55B & EP-425 & \textbf{7.2} & \textbf{4.6} & 13.6 & 14.9 & 43.2 & 16.7 & 101.4 & 28.2 \\
& WhisperLv3-X & 1.55B & EP-425 & \underline{7.4} & \underline{4.7} & \textbf{11.3} & \textbf{11.2} & 27.9 & \textbf{12.5} & 24.0 & 27.2 \\ \midrule
%\noalign{\vskip 0.5ex} \hdashline \noalign{\vskip 0.5ex}

\multirow{4}{*}[0pt]{TFS} & EBranch & 114M & EP-425 & 9.4 & 6.5 & 18.0 & 16.5 & 35.4 & 17.2 & 36.3 & 59.0 \\
& EBranch-XLSR & 114M\textsuperscript{¶} & EP-425 & 9.6 & 6.5 & 16.7 & 18.2 & 29.3 & 16.1 & 27.6 & 48.1 \\
& EBranch-w2vBERT2 & 114M\textsuperscript{¶} & EP-425 & 8.3 & 5.4 & 16.0 & 14.9 & \underline{27.2} & 14.4 & 26.7 & 42.4 \\
& \quad + 4-gram LM & 114M\textsuperscript{¶} & EP-425 & 8.0 & 5.4 & 15.6 & 13.4 & \textbf{27.1} & 13.9 & 26.6 & 41.9 \\
\midrule\midrule
\multirow{2}{*}[0pt]{BP} & WhisperLv3-X & 1.55B & BP-417 & 17.2 & 13.8 & 24.1 & 20.8 & 46.6 & 24.5 & 29.4 & \textbf{18.8} \\
& EBranch-w2vBERT2 & 114M\textsuperscript{¶} & BP-417 & 37.9 & 32.6 & 42.2 & 40.0 & 54.9 & 41.5 & 38.3 & 21.3 \\
\midrule

%Phi-4-MI & - & BP-417 & - & - & - & - & - & - & - & - & - \\
\multirow{2}{*}[0pt]{AAP} & WhisperLv3-X & 1.55B & (EP-425) $\rightarrow$ AAP-8 & 8.2 & 5.9 & \underline{12.1} & 11.7 & 28.9 & 13.4 & \textbf{22.7} & 26.3\\
& EBranch-w2vBERT2 & 114M\textsuperscript{¶} & (EP-425) $\rightarrow$ AAP-8 & 8.3 & 5.5 & 16.0 & 13.7 & 27.4 & 14.2 & 26.3 & 41.9 \\

\midrule
%Phi-4-MI & - & EP-425 + BP-417 & - & - & - & - & - & - & - & - & - \\
\multirow{2}{*}[0pt]{PT-All} & WhisperLv3-X & 1.55B & EP-425 + BP-417 + AAP-8 & 7.9 & \underline{4.7} & 12.3 & \underline{11.6} & 28.6 & \underline{13.0} & \underline{23.3} & \textbf{18.8} \\
& EBranch-w2vBERT2 & 114M\textsuperscript{¶} & EP-425 + BP-417 + AAP-8 & 8.7 & 5.5 & 16.4 & 13.5 & 28.0 & 14.4 & 24.6 & \underline{20.7} \\
\bottomrule
\end{tabular}
}
\vspace{-0.3cm}
\end{table*}
\vspace{-0.1cm}
\section{Experimental setup}
\label{sec:exp}
%As described in the previous section, we evaluate our proposed benchmark using several state-of-the-art supervised ASR foundation models:
The supervised foundation models described in the previous section %(i.e., WhisperLv3, OWSM-CTC v4, MMS-all, SeamlessM4T-v2 and Phi-4-MI.), 
were used through the Hugging Face platform\footnote{\url{https://huggingface.co}}, employing either zero-shot inference or fine-tuning approaches. For zero-shot inference, all models were used with their default configurations. Preliminary experiments indicated that WhisperLv3 and Phi-4-MI delivered the strongest performance, having thus been selected for fine-tuning on the EP training corpora. The full WhisperLv3 was fine-tuned for 10 epochs with a batch size of 64 and a gradient accumulation factor of 4, using a learning rate of 1e-5; Phi-4-MI was fine-tuned for 3 epochs with a batch size of 16 and the same learning rate. Only the audio components in Phi-4-MI were fine-tuned. For both models, 10\% of the initial steps were used for warm-up with a cosine learning rate scheduler. These settings were used for all fine-tuning experiments unless stated otherwise.          

All EBranch models were trained and evaluated using the ESPnet2 toolkit \cite{watanabe2018espnet}. We used the s3prl framework~\cite{s3prl}, which is natively integrated in ESPnet2, to incorporate SSL-based speech encoders as feature extractors for EBranch. Whereas w2v-BERT2 requires manual integration, XLSR is natively supported within the s3prl framework.
%To integrate upstream SSL models -- specifically the w2v-BERT2 speech encoder -- as a feature extractor for EBranch, we used the s3prl framework~\cite{s3prl}, which is natively integrated in ESPnet2. Whereas w2v-BERT2 requires manual integration, XLSR is natively supported within the s3prl framework.

Training of EBranch models followed the ESPnet LibriSpeech recipe, which combines an SSL model as a feature extractor with a conformer-based ASR architecture\footnote{\url{https://huggingface.co/espnet/simpleoier_librispeech_asr_train_asr_conformer7_wavlm_large_raw_en_bpe5000_sp}}. 
The encoder was adapted from the original recipe and comprises 12 layers. The decoder is a 6-layer Transformer derived from the same recipe. For both the encoder and decoder modules, we used Rotary Positional Embeddings (RoPE)~\cite{su2023roformerenhancedtransformerrotary}, which have demonstrated equal or superior performance in ASR tasks compared to absolute and relative positional encodings \cite{zhang2025benchmarking, longlibriheavy}. However, RoPE introduces instabilities in training convergence \cite{zhang2025benchmarking, longlibriheavy}. To mitigate this, we adopted a piecewise-linear learning rate schedule~\cite{owsm_ctc}, gradually increasing the learning rate as in \cite{longlibriheavy}: first to 2.0e-4 over the initial 15k steps, then to 2.0e-3 over the next 30k steps. All EBranch models were trained for 35 epochs using 13M batch bins. The resulting EBranch model comprises $\sim$114M parameters in total; this configuration was used for all training scenarios. To assess the full potential of our best-performing EBranch model for EP, we added a 4-gram language model (LM) at inference time. The LM was trained on the combined texts of the Europarl \cite{koehn2005europarl} and OpenSubtitles  \cite{opensubtitles2016} EP text-only corpora using the KenLM toolkit~\cite{heafield2011kenlm}.

All models use the same text normalizer, based on the standard normalization procedures used in Whisper. Performances are reported using word error rate (WER), and utterances longer than 30 seconds are excluded during training and fine-tuning. For consistency, all the experiments were conducted using a single NVIDIA A100 80GB GPU.      

%Note that the EBranch configuration remains consistent across both training scenarios -- whether using only Fbanks or incorporating a frozen SSL feature extractor. 

\vspace{-0.1cm}
\section{Results}
%\vspace{-0.05cm}
\label{sec:results}
%In the first three sections %CB: changed to follwoing, as it was unclear if itw as the first three of the paper
%The first three subsections focus on EP, the main target of the CAMÕES benchmark. We then conclude with results for BP and AAP varieties. %Dataset-specific results can be found on the CAMÕES HuggingFace webpage.
%\vspace{-0.05cm}
\subsection{Zero-shot foundation models}
%\vspace{-0.05cm}
The zero-shot results for the models in Section~\ref{CAMOES_framework}, covering the five CAMÕES EP domains mentioned in Section~\ref{CAMOES_benchmark}, are labelled ‘0-shot’ in Table~\ref{tab:results-all}.%, as well as the AAP and BP varieties. 

Among all the evaluated models, WhisperLv3-X achieves the best overall zero-shot performance for EP with a WER of 19.2\%. While WhisperLv3 performs well on the BN and T/L domains, it presents notable hallucinations in the RS and SI domains. These issues are greatly reduced %with the introduction of 
by %CB: to avoid widow line
WhisperLv3-X, with absolute WER improvements of over 10\% for both domains. This supports the claim that Whisper has a large tendency to hallucinate, in part due to the noise present in non-speech segments and to sample duration~\cite{whisper_hallucinations}.

The second-best performance for EP is achieved by Phi-4-MI with an average WER of 21.7\%; it also achieves the best performance in the RS domain with a WER of 15.5\%. It is worth noting that, in preliminary experiments, we evaluated three different prompts for transcribing EP audio with Phi-4-MI: 1) \textit{``Transcribe the audio clip into text''}, 2) \textit{``Transcribe the Portuguese audio clip into text''}, and 3) \textit{``Transcribe the European Portuguese audio clip into text''}. 
% CB: removed this paragraph because we are talking about the same thing%
The most specific prompt, which explicitly references EP, yielded the best results. 
%Similar preliminary experiments for AAP and BP identified optimal prompts as ``Transcribe the Accented Portuguese noisy audio clip into text." and ``Transcribe the Brazilian Portuguese noisy audio clip into text.", respectively. 
This suggests that prompt tuning has a measurable impact on zero-shot ASR performance, particularly in language- and dialect-sensitive contexts. Although not the focus of this work, to the best of our knowledge, we are among the first to use Phi-4-MI and explore prompting strategies for low-resource zero-shot ASR, opening new avenues for research. Contrarily, applying the most specific prompt to WhisperLv3-X did not yield any improvement; in fact, it led to performance degradation. This indicates that the Phi-4-MI LLM decoder is more responsive to prompt tuning than Whisper, especially in zero-shot ASR scenarios.   

MMS-all presents the worst %lowest % CB: replaced lowest by worst because low is good :)
result, possibly due to the small amount of Portuguese data used in its training (although the total number of hours is unknown). Notably, OWSM-CTC v4 is able to perform on par with SeamlessM4T-v2, despite its smaller size and having been trained with fewer data.

Overall, the top-performing zero-shot models -- \textbf{WhisperLv3-X} and \textbf{Phi-4-MI} -- were also those trained on the largest datasets (5M and 2.3M hours, respectively). Given their strong zero-shot performance and extensive pre-training, we selected these two models for fine-tuning. 

\vspace{-0.05cm}
\subsection{Fine-tuned and trained from scratch models}
\vspace{-0.05cm}
%Results regarding the fine-tuning (FT) of WhisperLv3 (and WhisperLv3-X) and Phi-4-MI, and those obtained with the EBranch models trained from scratch (TFS), can be found in the second and third blocks of Table \ref{tab:results-all}, respectively.
Table \ref{tab:results-all} shows the results for fine-tuned (FT) WhisperLv3, WhisperLv3-X, and Phi-4-MI models, as well as those obtained with the EBranch models trained from scratch (TFS).

The results show drastic performance improvements for the Whisper-type fine-tuned models, compared to those in the previous section.
%Considering the Whisper-type fine-tuned models, performance improves drastically for both versions. 
WhisperLv3-X achieves the lowest WER, with a relative improvement of 35\% compared to its zero-shot performance. Moreover, the fine-tuned WhisperLv3 model
presents fewer hallucination problems than the original model. Phi-4-MI obtains mixed results, with improved performance compared to the zero-shot version in the RS, BN and T/L domains, but higher WERs for the more demanding CS and SI domains. This highlights the difficulty of fine-tuning a multimodal LLM with data from a single modality.
%Comparatively, the performance gains achieved for Phi-4-MI are much lower -- $\sim$20\% relative improvement -- highlighting the difficulty of fine-tuning a multimodal LLM with data from a single modality.
Still, it is important to note that, to our knowledge, our work conducts the first evaluation and fine-tuning of Phi-4-MI -- a non-speech-centric multimodal LLM -- for low-resource ASR.

Regarding the EBranch models trained from scratch, we find that using more powerful SSL foundation models (i.e., trained with more data) as speech encoders provides large improvements over the baseline FBank features, as expected. We further observe that the w2v-BERT2 encoder (Section~\ref{CAMOES_framework}) outperforms the XLSR encoder, despite having only 27\% of its parameters. However, XLSR was pre-trained on just 436k hours—only 9.7\% of the data used to pre-train w2v-BERT2.
%Moreover, although the best version of EBranch (EBranch-w2vBERT2 + 4 gram LM) is not able to achieve the same performance as WhisperLv3-X, the absolute difference between the two models is only $\sim1\%$. In addition, to achieve this performance Whisper requires VAD (-X extension), and, without it, the EBranch-SSL is able to outperform it. 
Moreover, although the best version trained from scratch (EBranch-w2vBERT2 + 4-gram LM) does not achieve the performance of WhisperLv3-X fine-tuned --  the absolute difference is less than 2\% (Avg.) -- it largely surpasses the performance of the fine-tuned Whisper without the VAD pre-processing decoding strategy. This is a strong result for a model with 114M trainable parameters trained from scratch.

\begin{figure*}[t]
    \centering
    \includegraphics[width=0.79\linewidth]{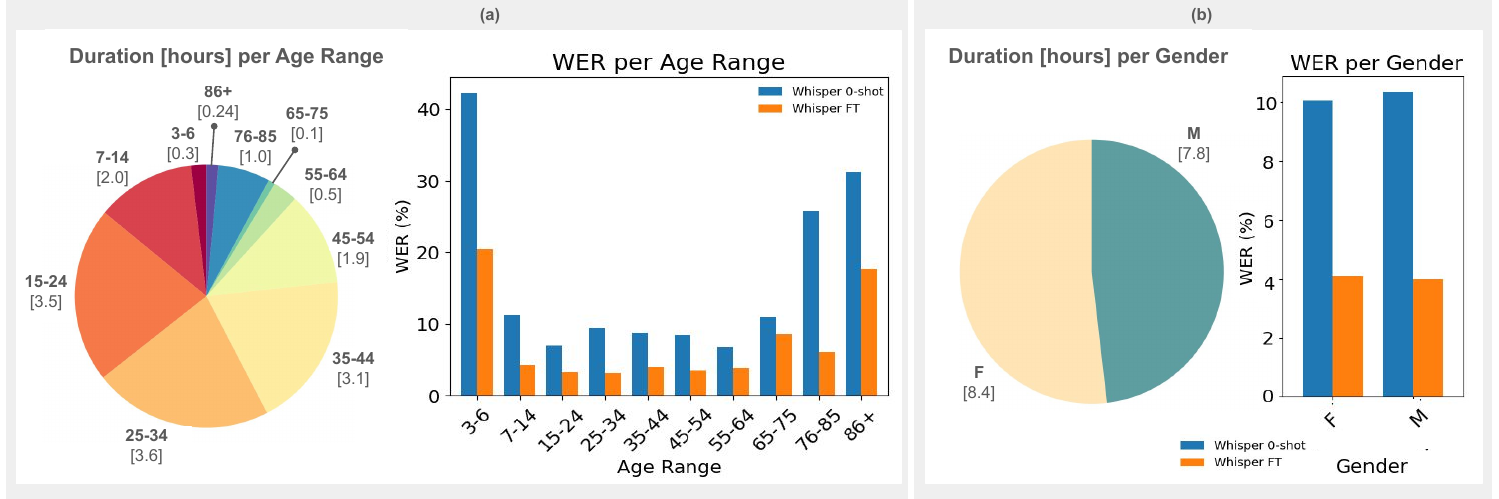}
    \vspace{-6pt}
    \caption{WER [\%] performance for \textit{WhisperLv3-X} zero-shot and fine-tuned, per age range (a) and per gender (b), on the \textbf{RS} domain (BD-Publico, PT\_Adults, PT\_Children, PT\_Elderly, and SpeechDat). The figure also shows the number of hours per age range and gender considered for this analysis.}
    \vspace{-12pt}
    \label{fig:demographics_wer}
\end{figure*}

\vspace{-0.05cm}
\subsection{Demographic analysis} % tentative sub-section name
%\vspace{-0.05cm}
Figure \ref{fig:demographics_wer} shows the results obtained for the best foundation model, WhisperLv3-X, with and without finetuning, for a subset of the \textbf{RS} domain as an illustrative example. The results are divided by age range (a) and gender (b), to understand different model behaviours across demographic groups. %In Fig. \ref{fig:demographics_wer} (a), we observe that the model has the most trouble recognizing the speech of very young children (aged between 3-6) and elders aged above 86. 
%%This result is intuitive, since children in this age range have not yet fully developed their speaking ability, whereas elders aged above 86 are likely to suffer from impaired speech due to age-related health conditions \cite{}. 
%Across all other age groups the model presents a relatively stable performance, showing only slight variations.
%%For the age group between 76-85 there is a large improvement from the zero-shot to the fine-tuned model, likely due to the fact that the fine-tuning dataset contains individuals in this, and in adjacent, age ranges.
%
%CB: Suggestion of rewrite:
Fig.~\ref{fig:demographics_wer}(a) shows that the model struggles most with speech from very young children (ages 3–6) and elderly speakers (86+), while performance across other age groups remains relatively stable. For gender, we observe in Fig.~\ref{fig:demographics_wer}(b) a very balanced performance for male and female speakers for both versions of the model. 
%the zero-shot model, a difference which is reduced even further after fine-tuning. %Overall, these results indicate that our model is largely without biases for the considered demographics.
% CB: suggestion
Overall, the fine-tuned model outperforms the zero-shot baseline across all demographic groups, suggesting reduced bias and overall improved robustness.

%-- Comparison by gender (bias) and age group?
%, with specific focus on Portuguese. 
%Notes for the discussion:
%- In all domains we improved for whisperX
%-- Notably, 
%\subsection{Models trained from scratch} %tentative subsection title if even sub-section is necessary

\vspace{-0.05cm}
\subsection{Brazilian, African and Asian Portuguese Varieties} 
%\vspace{-0.05cm}
%tentative sub-section name
% \subsection{Brazilian and other Portuguese Varieties}

%\input{tables/check}

%and BP, with WERs of 19.2\% and 24.5\%, respectively. 
%For AAP, the best-performing model is Phi-4-MI, with an average WER of 27.1\%, outperforming WhisperLv3-X, which achieves 29.0\%.  
%, while for BP, WhisperLv3 ranks second, with a WER of 24.6\%. 

%Notes for the discussion:
%We suspect the 0-shot models have seen a lot of BP. and specifically bp from specific datasets that are in our benchmark 
%It was expected to lose performance in BP

%{\color{red}!! Notes: Os africanos fazem um grau de redução vocálica intermédio entre nós e os brasileiros
%(Isabel vai ver isto)}

Regarding the zero-shot evaluation, Table~\ref{tab:results-all} shows that Phi-4-MI achieves the best performance on AAP with a WER of 27.1\%, while WhisperLv3-X performs best on BP with a WER of 24.6\%. As with EP, preliminary prompt engineering for Phi-4-MI identified the optimal prompts as \textit{``Transcribe the Accented Portuguese noisy audio clip into text''} for AAP and \textit{``Transcribe the Brazilian Portuguese noisy audio clip into text''} for BP. 
%
%Furthermore, across all zero-shot foundation models, we observe that performance on BP is consistently stronger than on AAP, suggesting that these models are exposed to more BP data during pretraining. 
%CB suggestion:
%Furthermore performance on BP consistently surpasses that on AAP varieties across all zero-shot foundation models, suggesting a greater exposure of the models to BP during pre-training. {\color{red}Similarly, when comparing BP and EP (within the TV and SI domains for a fair domain-level comparison), models again perform better on BP. These results support the hypothesis that foundational models have been trained on more BP than either AAP or EP speech data.}
% CB second suggestion
We refrain from making direct column-wise comparisons across language varieties, as the datasets differ in tasks, recording conditions, and other factors. However, when comparing similar domains -- specifically, the average performance on the CS and SI domains in EP against those in BP and AAP -- zero-shot models tend to perform better on BP. This suggests greater exposure to BP data during pre-training.

%Fine-tuning these models on EP has mixed results when it comes to AAP and BP: WhisperLv3-X fine-tuned on EP shows an improvement of 5\% WER for AAP, while for BP performance is always degraded. This further supports the claim that foundation models have mostly been exposed to the Brazilian variety of Portuguese.

%EBranch models trained on EP from scratch also have mixed performances for AAP and BP. For BP in particular, the performance of all of these models is worse than the zero-shot foundation models. Nonetheless, there is a clear positive contribution of the SSL feature extractors on the performance of the EBranch models on AAP and BP. 

Fine-tuning with EP data yields mixed results for AAP and BP. Performance improves $\sim$5\% WER for AAP with the fine-tuned WhisperLv3-X model, but consistently degrades for BP -- reinforcing the notion that foundation models are primarily trained on BP.
%Fine-tuned WhisperLv3-X improves AAP results by 5\% WER, but consistently degrades performance on BP --reinforcing the notion that foundation models are primarily trained on BP. 
%Similarly, EBranch models trained from scratch on EP show variable performance across varieties. For BP, their performance lags behind that of zero-shot foundation models. 
% CB rewrite to try to save widow line:
%{\color{red}Similarly, EBranch models trained from scratch with EP speech show variable performance across varieties, with BP performance trailing that of zero-shot models}. 
Similarly, EBranch models trained from scratch with EP speech also show improvements for AAP, with results degrading for BP, when compared to zero-shot models.
However, the use of SSL feature extractors clearly benefits the performance of EBranch models on both AAP and BP, compared to the FBank baseline. 

%To obtain the strongest possible model for AAP and BP we selected the best-performing models for EP -- WhisperLv3-X and EBranch-w2vBERT2 -- and finetuned/trained these models on in-domain data for each of these accent groups. In the end we trained/fine-tuned with all data together (EP-425 + BP-417 + AAP-8).
Different approaches were followed to build stronger variety-dependent models. For BP, we fine-tuned WhisperLv3-X and trained an EBranch-w2vBERT2 model from scratch, similarly to what was done for EP. For AAP, given the limited size of the training set (8h), we opted to fine-tune the best performing models trained on EP speech (WhisperLv3-X and EBranch-w2vBERT2).
%and given the limited size of the AAP training set (8h), for AAP we opted to fine-tune the best performing models trained on EP  (WhisperLv3-X and EBranch-w2vBERT2). In contrast, for BP we followed a similar approach as with EP, fine-tuning WhisperLv3-X and training from scratch EBranch-w2vBERT2. 
The best performance for AAP was achieved by WhisperLv3-X pre-fine-tuned on EP, reaching a WER of 22.7\% -- a relative improvement of 20\% over the best zero-shot result. This is a good gain given the small amount of fine-tuning data. Nonetheless, the impact of AAP-8 is limited by its lack of coverage of test-set varieties like Macao, Goa, and East Timor. For BP, fine-tuning on BP-417 yielded WhisperLv3-X as the best model, achieving 18.8\% WER, with EBranch-w2vBERT2 close behind at 21.3\% WER. %In contrast to EP, the performance gains for BP over zero-shot models were modest, despite the comparable dataset sizes. This may be due to the fact that strong foundation models have already seen significant amounts of BP data during pre-training, limiting the impact of additional fine-tuning. Additionally, the EP training data spans more diverse domains, which may have helped models generalize more effectively.

An interesting characteristic of these results is that models fine-tuned on one variety of Portuguese tend to perform poorly on the other. For example, the best EP model -- WhisperLv3-X fine-tuned on EP-425 -- achieves a strong WER of 12.5\% on EP but degrades to 27.2\% on BP. Conversely, the best BP model -- WhisperLv3-X trained on BP-417 -- achieves 18.8\% WER on BP but only 24.5\% on EP. To address this issue, we also fine-tuned WhisperLv3-X and trained an EBranch-w2vBERT2 model on the whole multi-variety corpus EP-425+BP-417+AAP-8 (denoted as PT-All), to assess cross-varietal robustness.
Table~\ref{tab:results-all} shows that both the fine-tuned WhisperLv3-X and the EBranch-w2vBERT2 models trained on PT-All achieve performances
%across the EP, BP and AAP varieties 
comparable to their variety-specific fine-tuned/TFS versions across all varieties. Notably, the EBranch-w2vBERT2 architecture achieves its best average results on BP and AAP, while preserving its performance on EP. These findings suggest that joint training with multi-varietal speech helps mitigate the performance asymmetry previously observed between EP and BP with variety-specific models, and improves the models' ability to generalize. 
%generalization capability of the models. 
More importantly, this approach yields a single model that achieves SOTA results for Portuguese ASR across all varieties.
%For models trained jointly on EP-425, BP-417, and AAP-8 (last two rows of Table~\ref{tab:results-all}), we observe that both WhisperLv3-X and EBranch-w2vBERT2 maintain performance comparable to their respective individually fine-tuned/trained versions across EP, BP, and AAP.

%Finally, we evaluate SOTA performance on BP 
Finally, we compare our models against the current SOTA for BP using the test sets of CORAA and MuPe (Section ~\ref{sec:data_BR}).
%, introduced in Section~\ref{sec:data}. 
%MuPe, a challenging new corpus of 289 life-story interviews, features diverse speakers across age, education, and regional accents. 
As shown in Table~\ref{tab:bp_focused}, on average, our models outperform prior work. % with the results for MuPe having been achieved in a zero-shot setting. For instance, 
The WhisperLv3-X model from PT-All on MuPe achieves a WER just 2.6\% higher than the 15.9\% of Distill-WhisperLv3 which was fine-tuned on this dataset, whereas our models were not. Moreover, the PT-All models perform on par with the BP-only models, highlighting a strong generalization of our approach across Portuguese varieties.     
\begin{table}[t]
\setlength{\tabcolsep}{2pt}
\caption{WER [\%] for prior BP SOTA vs. our BP-only and PT-All models.}

\vspace{-6pt}
\centering
\resizebox{0.35\textwidth}{!}{
\label{tab:bp_focused}
\begin{tabular}{l|c|c|c|c}
\toprule
\multicolumn{1}{c|}{\bf Model} & \bf Training Data &\bf CORAA & \bf MuPe & \bf Average \\
\midrule
XLSR53-CTC\cite{conneau2020unsupervised,candido2023coraa} & - & 24.2 & 28.8 & 26.5 \\
 Distil-WhisperLv3 \cite{leal2025mupe} & - & 26.1 & \bf 15.9 & 21.1 \\
\midrule
\midrule
WhisperLv3-X & BP-417 &\underline{14.5} & \underline{18.1} & \bf 16.3 \\
EBranch-w2vBERT2 & BP-417 & 17.5 & 20.2 & 18.9 \\

WhisperLv3-X & PT-All & \bf{14.0} & 18.5 & \bf 16.3 \\
EBranch-w2vBERT2 & PT-All & 17.3 & 19.1 & \underline{18.2} \\
\bottomrule
\end{tabular}}
\vspace{-16pt}
\end{table}

\vspace{-0.05cm}
\section{Conclusions}
\vspace{-0.05cm}
\label{sec:conclusions}
%Future work: prompts/tags during training with varieties
%-- + punctuation
%-- weakly supervised approaches + model distillation

This work introduces CAMÕES --the first comprehensive evaluation benchmark for EP, covering a broad range of age groups and domains, as well as other Portuguese varieties, including AAP and BP%We evaluate and fine-tune strong speech-centric foundation models such as WhisperLv3 and OWSM-CTC v4, alongside a multimodal LLM (Phi-4-MI).
--, and establishes a new SOTA reference for EP and AAP.
We evaluate a range of speech-centric foundation models, including WhisperLv3, MMS-all, SeamlessM4T-v2 and OWSM-CTC v4, as well as a multimodal LLM (Phi-4-MI), and fine-tune the strongest candidates based on their zero-shot performance.
We also explore zero-shot prompt tuning with the Phi-4-MI LLM, showing its effectiveness. Additionally, we train EBranch-w2vBERT2 from scratch, achieving performances close to our best fine-tuned model, WhisperLv3-X. Joint training on EP, BP, and AAP matches the performances of variety-specific fine-tuned models, yielding a robust single model that generalizes well, with SOTA performance across Portuguese varieties.  
In future work, we plan to enhance our models by leveraging large-scale unlabeled data sources, such as VoxPopuli~\cite{wang2021voxpopuli}, and using weakly supervised learning~\cite{whisper} and knowledge distillation~\cite{udistilwhisper} techniques to improve model performance. Furthermore, we aim to explore online repositories as potential sources of new data.
%\section*{Acknowledgment}

%\newpage % ALBERTO: Just for controlling references an space

\bibliographystyle{ieeetr}
\bibliography{bibliography}~\footnote{© 2025 IEEE. Personal use of this material is permitted. Permission from IEEE must be obtained for all other uses, in any current or future media, including reprinting/republishing this material for advertising or promotional purposes, creating new collective works, for resale or redistribution to servers or lists, or reuse of any copyrighted component of this work in other works.}

\vspace{12pt}
%\color{red}
%IEEE conference templates contain guidance text for composing and formatting conference papers. Please ensure that all template text is removed from your conference paper prior to submission to the conference. Failure to remove the template text from your paper may result in your paper not being published.

\end{document}